\documentclass[10pt,twocolumn,letterpaper]{article}

\usepackage{iccv}
\usepackage{times}
\usepackage{epsfig}
\usepackage{graphicx}
\usepackage{amsmath}
\usepackage{amssymb}
\usepackage{paralist}
\usepackage{color}

\usepackage{setspace}
\usepackage[pagebackref=true,breaklinks=true,letterpaper=true,colorlinks,bookmarks=false,citecolor=blue,linkcolor=blue]{hyperref}

\iccvfinalcopy 
\usepackage{verbatim}
\def\iccvPaperID{211} 

\ificcvfinal\fi

\begin{document}
\title {SegFlow: Joint Learning for Video Object Segmentation and Optical Flow}

	\author{Jingchun Cheng$^{1,2}$
    \hspace{0.15in} Yi-Hsuan Tsai$^{2}$ 
	\hspace{0.15in} Shengjin Wang$^{1}$
	\hspace{0.15in} Ming-Hsuan Yang$^{2}$
	\vspace{2mm} \\
	\hspace{0.15in} $^{1}$Tsinghua University \hspace{0.15in} $^{2}$University of California, Merced \\
	\hspace{0.1in} $^{1}${\tt\small chengjingchun@gmail.com, wgsgj@tsinghua.edu.cn}
	 \hspace{0.3in} $^{2}${\tt\small \{ytsai2, mhyang\}@ucmerced.edu}
	}

\maketitle
\thispagestyle{empty}

\begin{abstract}
This paper proposes an end-to-end trainable network, SegFlow, for simultaneously predicting pixel-wise object segmentation and optical flow in videos.
The proposed SegFlow has two branches where
useful information of object segmentation and optical
flow is propagated bi-directionally in a unified framework.
The segmentation branch is based on a fully convolutional network, which has been proved effective in image segmentation task,
and the optical flow branch takes advantage of the FlowNet model.
The unified framework is trained iteratively offline to learn a generic notion, and fine-tuned online for specific objects.
Extensive experiments on both the video object segmentation and optical flow datasets demonstrate that introducing optical flow improves the performance of segmentation and vice versa, against the state-of-the-art algorithms.
\end{abstract}


\section{Introduction}
\begin{figure}[t]
	\begin{center}
    \includegraphics[width=1.0\linewidth]{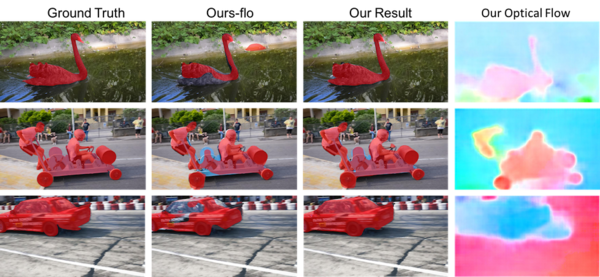}
	\end{center}
	\caption{
		An illustration of the main idea in the proposed \textit{SegFlow} model.
		Our model produces better segmentation results than the one without using the optical flow (Ours-flo), where the flow within the object is smooth and complete, providing a guidance to improve segmentation outputs.
		}
	\label{fig:intro}
	\vspace{4mm}
\end{figure}

Video analysis has attracted much attention in recent years due to the
numerous vision applications, such as autonomous driving \cite{geiger2012we, chen2015deepdriving, ros2015vision}, video surveillance \cite{tian2005robust, cohen1999detecting, junejo2004multi} and virtual reality \cite{anderson2016jump}.
To understand the video contents for vision tasks, it is essential to know the object status (e.g., location and segmentation) and motion information (e.g., optical flow).
In this paper, we address these problems simultaneously, i.e., video object segmentation and optical flow estimation, in which these two tasks have
been known to be closely related to each other \cite{tsai2016video, sevilla2016optical}.
Figure \ref{fig:intro} illustrates the main idea of this paper.

For video object segmentation \cite{marki2016bilateral}, it assumes that the object mask is known in the first frame, and the goal is to assign pixel-wise foreground/background labels through the entire video.
To maintain temporally connected object segmentation, optical flow is typically used
to improve the smoothness across the time \cite{Papazoglou_ICCV_2013}.
However, flow estimation itself is a challenging problem and is often inaccurate,
and thus the provided information does not always help segmentation.
For instance, when an object moves fast, the optical flow methods \cite{Baker_ICCV_2007, Brox_PAMI_2011, Sun_IJCV_2014} are not effective in capturing the movement and hence generate incomplete flow within the object (see Figure \ref{fig:DAVIS_flow_compare} for an example).
To overcome this problem, bringing the objectness information (i.e., segmentation) can guide the algorithm to determine where the flow should be smooth (within the object).
A few algorithms have been developed
to leverage both information from the objectness and motion discussed above.
In \cite{tsai2016video}, a method is proposed to simultaneously perform
object segmentation and flow estimation, and then updates both results iteratively.
However, the entire process is optimized online and is time-consuming, which limits the applicability to others tasks.

Based on the above observations, we propose a learning-based approach to jointly predict object segmentation and optical flow in videos, which allows efficient inference during testing.
We design a unified, end-to-end trainable convolutional neural network (CNN), which we refer to as the \textit{SegFlow}, that contains one branch for object segmentation and another one for optical flow.
For each branch, we learn the feature representations for each task, where the segmentation branch focuses on the objectness and the optical flow one exploits the motion information.
To bridge two branches to help each other, we propagate the learned feature representations bi-directionally.
As such, these features from one branch can facilitate the other branch while obtaining useful gradient information during back-propagation.

One contribution of the proposed network is the bi-directional architecture that enables the communication between two branches, whenever the two objectives of the branches are closely related and can be jointly optimized.
To train this joint network, a large dataset with both ground truths of two tasks (i.e., foreground segmentation and optical flow in this paper) is required.
However, such dataset may not exist or is difficult to construct.
To relax such constrains, we develop an iterative training strategy that only requires one of the ground truths at a time, so that the target function can still be optimized and converge to a solution where both tasks achieve reasonable results.

To evaluate our proposed network, we carry out extensive experiments on both the video object segmentation and optical flow datasets.
%
%
We compare results on the DAVIS segmentation benchmark \cite{Perazzi2016} with or without providing motion information, and evaluate the optical flow performance on the Sintel \cite{Butler:ECCV:2012}, Flying Chairs \cite{fischer2015flownet} and Scene Flow \cite{MIFDB16} datasets.
In addition, analysis on the network convergence is presented to demonstrate our training strategy.
We show that the bi-directional network through feature propagation performs favorably against state-of-the-art algorithms on both video object segmentation and optical flow tasks in terms of visual quality and accuracy.

The contributions of this work are summarized below:
\begin {compactitem}
\item We propose an end-to-end trainable framework for simultaneously predicting
pixel-wise foreground object segmentation and optical flow in videos.

\item We demonstrate that optical flow and video object segmentation tasks are complementary, and can help each other through feature propagation in a bi-directional framework.

\item We develop a method to train the proposed joint model without the need of a dataset that contains both segmentation and optical flow ground truths.

\end {compactitem}

\begin{figure*}
	\begin{center}
		\includegraphics[width=0.9\linewidth]{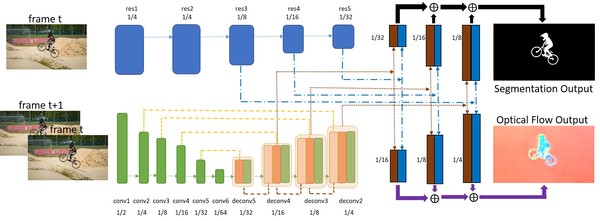}
	\end{center}
	\vspace{-3mm}
	\caption{
		The proposed \textit{SegFlow} architecture.
		Our model consists of two branches, the segmentation network based on a fully-convolutional ResNet-101 and the flow branch using the FlowNetS \cite{fischer2015flownet} structure.
		In order to construct communications between two branches, we design an architecture that bridges two networks during the up-sampling stage.
		Specifically, feature maps are propagated bi-directionally through concatenations at different scales with proper operations (i.e., up-sampling or down-sampling) to match the size of different features.
		Then an iterative training scheme is adopted to jointly optimize the loss functions for both segmentation and optical flow tasks.
		}
	\label{fig:framework}
	\vspace{-3mm}
\end{figure*}
%
\section{Related Work}
%
{\flushleft {\bf Unsupervised Video Object Segmentation.}}
Unsupervised methods aim to segment foreground objects without any knowledge of the object (e.g., an initial object mask).
Several methods have been proposed to generate object segmentation via 
\cite{rahtu2010segmenting, faktor2014video, Wang_CVPR_2015}, optical flow \cite{Brox_ECCV_2010, Papazoglou_ICCV_2013} or superpixel \cite{grundmann2010efficient, xu2012streaming, galasso2012video}.
To incorporate higher level information such as objectness, object proposals are used  to track object segments and
generate consistent regions through the video \cite{Lee_ICCV_2011, Li_ICCV_2013}.
However, these methods usually require heavy computational loads to generate region proposals and associate thousands of segments, making such methods only applicable to offline applications.
%
{\flushleft {\bf Semi-supervised Video Object Segmentation.}}
Semi-supervised methods \cite{Galasso_ICCV_2013, Zhang_CVPR_2013, Nagaraja_ICCV_2015} assume an object mask in the first frame is known, and track this object mask through the video.
To achieve this, existing approaches focus on propagating superpixels \cite{Jain_ECCV_2014}, constructing graphical models \cite{marki2016bilateral, tsai2016video} or utilizing object proposals \cite{perazzi2015fully}.
Recently, CNN based methods \cite{khoreva2016learning, caelles2016one} are developed by combining offline and online training processes on static images.
Although significant performance has been achieved,
the segmentation results are not guaranteed to be smooth in the temporal domain.
In this paper, we use CNNs to jointly estimate optical flow and provide the learned motion representations to generate consistent segmentations across time.

{\flushleft {\bf Optical Flow.}}
It is common to apply optical flow to video object segmentation to maintain motion consistency.
One category of the approaches is to solve a variational energy minimization problem \cite{Baker_ICCV_2007, Brox_PAMI_2011, Sun_IJCV_2014} in a coarse-to-fine scheme.
To better determine the correspondences between images, matching based optimization algorithms \cite{weinzaepfel2013deepflow, revaud2015epicflow} are developed, in which these methods usually require longer processing time.
On the other hand, learning based methods are more efficient, which can be achieved via Gaussian mixture models \cite{rosenbaum2013learning}, principle components \cite{Wulff_CVPR_2015} or convolutional networks \cite{fischer2015flownet, MIFDB16}.
Considering the efficiency and accuracy, we apply the FlowNet \cite{fischer2015flownet} as our baseline in this work, while we propose to improve optical flow results by feeding the information from the segmentation network as guidance, which is not studied by the above approaches.
%

{\flushleft {\bf Fusion Methods.}}
The joint problem of video segmentation and flow estimation has been
studied by layered models \cite{Chang_ICCV_2013, Sun_CVPR_2013}.
Nevertheless, such methods rely on complicated optimization during inference, thereby limiting their applications.
Recently, significant efforts have been made along the direction of video object segmentation while considering optical flow.
In \cite{khoreva2016learning}, a network that uses pre-computed optical flow as an additional input to improve segmentation results is developed.
Different from this work, our model only requires images as the input, and we aim to jointly learn useful motion representations to help segmentation.

Closest in scope to our work is the ObjectFlow algorithm (OFL) \cite{tsai2016video}
that formulates an objective function to iteratively optimize segmentation and optical flow energy functions.
However, this method is optimized online and is thus computationally expensive. In addition, it requires the segmentation results before updating the estimation for optical flow.
In contrast, we propose an end-to-end trainable framework for simultaneously predicting pixel-wise foreground object segmentation and optical flow.
%


%
\section{SegFlow}
Our goal is to segment objects in videos, as well as estimate the optical flow between frames.
Towards this end, we construct a unified model with two branches, a segmentation branch based on fully-convolutional network, and an optical flow branch based on
the FlowNetS \cite{fischer2015flownet}.

Due to the lack of datasets with both segmentation and optical flow annotations, we initialize the weights of two branches from legacy models trained on different datasets,
and optimize the \textit{SegFlow} on segmentation and optical flow datasets via iterative offline training and online finetuning.
In the following, we first introduce the baseline model of each the segmentation and optical flow branch, and explain how we construct the joint model using the proposed bi-directional architecture.
The overall architecture of our proposed joint model is illustrated in Figure \ref{fig:framework}.

\subsection{Segmentation Branch}
\label{sec:SegBranch}
Inspired by the effectiveness of fully-convolutional networks in image segmentation \cite{long2015fully} and the deep structure in image classification \cite{he2016deep, srivastava2015highway},
we construct our segmentation branch based on the ResNet-101 architecture \cite{he2016deep}, but modified for binary (foreground and background) segmentation predictions as follows:
1) the fully-connected layer for classification is removed, and
2) features of convolution modules in different levels are fused together for obtaining more details during up-sampling.

The ResNet-101 has five convolution modules, and each consists of several convolutional layers, Relu, skip links and pooling operations after the module.
Specifically, we draw feature maps from the 3-th to 5-th convolution modules after pooling operations, where score maps are with sizes of 1/8, 1/16, 1/32 of the input image size, respectively.
Then these score maps are up-sampled and summed together for predicting the final output (upper branch in Figure \ref{fig:framework}).

A pixel-wise cross-entropy loss with the softmax function $\mathbb{E}$ is used during optimization.
To overcome imbalanced pixel numbers between foreground and background regions, we use the weighted version as adopted in \cite{Xie_ICCV_2015}, and the loss function is defined as:
	\begin{align}
	\mathcal{L}_s(X_t) = - (1-w)\sum_{i,j\in fg} \log \mathbb{E}(y_{ij} = 1;\theta) \notag \\
	- w\sum_{i,j\in bg} \log \mathbb{E}(y_{ij} = 0;\theta),
	\label{eq:seg}
	\end{align}

\noindent
where $i, j$ denotes the pixel location of foreground $fg$ and background $bg$, $y_{ij}$ denotes the binary prediction of each pixel of the input image $X$ at frame $t$, and $w$ is computed as the foreground-background pixel-number ratio.
\subsection{Optical Flow Branch}
Considering the efficiency and accuracy, we choose the FlowNetS \cite{fischer2015flownet} as our baseline for flow estimation.
The optical flow branch uses an encoder-decoder architecture with additional skip links for feature fusions (feature concatenations between the encoder and decoder).
In addition, a down-scaling operation is used at each step of the encoder, where each step of the decoder up-samples back the output (see the lower branch in Figure \ref{fig:framework}).
Based on such structure, we find that it shares similar properties with the segmentation branch and their feature representations are in similar scales, which enables plausible connections to the segmentation model, and vice versa, where we will introduce in the next section.

To optimize the network, the optical flow branch uses an endpoint error (EPE) loss as adopted in \cite{fischer2015flownet}, which is defined as the following:
	\begin{equation}
	\mathcal{L}_f(X_t, X_{t+1}) = \sum_{i,j} (  (u_{ij} - \delta_{u_{ij}})^2 + (v_{ij} - \delta_{v_{ij}})^2 ),
	\label{eq:flow}
	\end{equation}
where $u_{ij}, v_{ij}$ denotes the motion at pixel $(i, j)$ of input images from $X_t$ to $X_{t+1}$, and $\delta_{u_{ij}}$ and $\delta_{v_{ij}}$ are network predictions.
We use the images at frame $t$ and $t+1$ as the computed optical flow should align with the segmentation output (e.g., object boundaries) at frame $t$, so that their information can be combined later naturally.
\subsection{Bi-directional Model}
\label{sec:joint_model}
In order to make communications between two branches as mentioned above, we propose a unified structure, \textit{SegFlow}, to jointly predict segmentation and optical flow outputs.
Therefore, the new optimization goal becomes to solve the following loss function that combines \eqref{eq:seg} and \eqref{eq:flow}:
$\mathcal{L}(X) = \mathcal{L}_s(X) + \lambda \mathcal{L}_f(X)$.
As shown in Figure \ref{fig:framework}, our architecture propagates feature maps between two branches bi-directionally at different scales for the final prediction.
For instance, features from each convolution module in the segmentation branch are first up-scaled (to match the size of optical flow features), and then concatenated to the optical flow branch.
Similar operations are adopted when propagating features from segmentation to flow.
Note that, a convolutional layer is also utilized (with the channel number equal to the output channel number) after fused features for network predictions, further regularizing the information from both the segmentation and optical flow branches.

Different from directly using final outputs to help both tasks \cite{tsai2016video}, we here utilize information in the feature space.
One reason is that our network is able to learn useful feature representations (e.g., objectness and motion) at different scales.
In addition, with the increased model capacity but without adding too much burden for training the network, the joint model learns better representations than the single branch. For instance, the single flow network does not have the ability to learn representations similar to the segmentation branch, while our model provides the chance for two tasks sharing their representations.
Note that, our bi-directional model is not limited to the current architecture or tasks, while it should be a generalized framework that can be applied to co-related tasks.
\section{Network Implementation and Training}
In this section, we present more details regarding how we train the proposed network.
To successfully train the joint model, a large-scale dataset with both the segmentation and optical flow ground truths is required.
However, it is not feasible to construct such a dataset.
Instead, we develop a training procedure that only needs one of the ground truths at a time by iteratively updating both branches and gradually optimizing the target function.
In addition, a data augmentation strategy is described for both tasks to enhance the diversity of data distribution and match the need of the proposed model.
%
%
%
\begin{figure}[t]
	\begin{center}
		\includegraphics[width=0.95\linewidth]{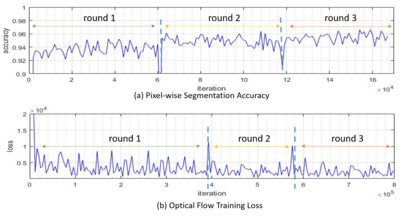}
	\end{center}
	\vspace{-3mm}
	\caption{
		During offline training, (a) shows the training accuracy for object segmentation, while (b) presents the loss for optical flow, with respect to the number of training iterations (both results are obtained on a training subset).
		After three rounds, convergences can be observed for both segmentation and optical flow.
	}
	\label{fig:accuracy_training}
	\vspace{-3mm}
\end{figure}
\subsection{Network Optimization}
First, we learn a generic model by iteratively updating both branches, where the goal of the segmentation network at this stage is to segment moving objects.
%
To focus on a certain object (using the mask in the first frame), we then finetune the model for the segmentation branch on each sequence of
the DAVIS dataset for online processing.
{\flushleft {\bf Iterative Offline Training.}}
%
%
To start training the joint model, we initialize two branches using the weights from ResNet-101 \cite{he2016deep} and FlowNetS \cite{fischer2015flownet}, respectively.
When optimizing the segmentation branch, we freeze the weights of the optical flow branch, and train the network on the DAVIS training set.
We use SGD optimizer with batch size 1 for training, starting from learning rate 1e-8 and decreasing it by half for every 10000 iterations.
%

For training the optical flow branch, similarly we fix the segmentation branch and only update the weights in the flow network using the target optical flow dataset (described in Section \ref{sec:dataset}).
To balance the weights between two different losses, we use a smaller learning rate 1e-9 for the EPE loss in \eqref{eq:flow}, addressing the $\lambda$ in the combined loss in Section \ref{sec:joint_model}.
Note that, to decide when to switch the training process to another branch, we randomly split a validation set and stop training the current branch when the error on the validation set reaches a convergence.
In addition, this validation set is also used to select the best model with respect to the iteration number \cite{fischer2015flownet}.

\begin{figure}
	\begin{center}
		\includegraphics[width=1.0\linewidth]{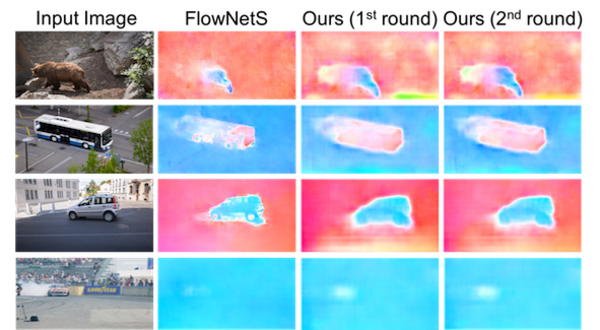}
	\end{center}
	\vspace{-1mm}
	\caption{
		Iteratively improving optical flow results on DAVIS.
		Given an input image, we show the flow estimation from the initial model, FlowNetS \cite{fischer2015flownet}, and our results during optimizing the \textit{SegFlow} in the first and the second round.
		The results are gradually improved during optimization.
	}
	\label{fig:DAVIS_flow_compare}
	\vspace{-3mm}
\end{figure}

For this iterative learning process, each time the network focuses on one task in a branch, while obtaining useful representations from another branch through feature propagation.
Then after switching to train another branch, better features learned from the previous stage are used in the branch currently optimized.
We show one example of how the network gradually move toward a convergence by iteratively training both branches in Figure \ref{fig:accuracy_training} (with three rounds).
In addition, Figure \ref{fig:DAVIS_flow_compare} shows visual improvements during iteratively updating the flow estimation.
{\flushleft {\bf Online Training for Segmentation.}}
The model trained offline is able to separate moving object from the video.
To adapt the model on a specific object for online processing, we finetune the segmentation network using the object mask in the first frame on each individual sequence.
Here, we call the process online in the semi-supervised setting, as the model is needed to update with the guidance of mask in the first frame before testing on the sequence.

%
Each mask is then augmented to multiple training samples for both branches to increase the data diversity (described in Section \ref{sec:DA}).
After data augmentation, we use the same training strategy introduced in the offline stage with a fixed learning rate of 1e-10.
At this stage, we note that the flow branch still provides motion representations to segmentation, but does not update the parameters.
%
%
\subsection{Data Augmentation}
\label{sec:DA}
\begin{figure}
	\begin{center}
		\includegraphics[width=1.0\linewidth]{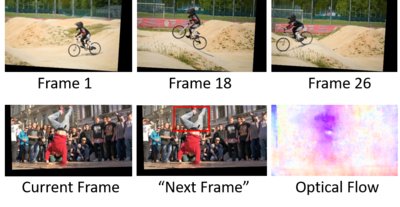}
	\end{center}
	\vspace{-5mm}
	\caption{Examples for data augmentation.
		The first row shows the data augmentation for segmentation with the same transform through the video for maintaining the temporal consistency.
		The second row presents one example of the augmented flow, where the transform is applied on the object mask to simulate the slight movement in the ``next frame'' (highlighted within the red rectangle), where the optical flow shows the corresponding transform.
	}
	\label{fig:data_augmentation}
	\vspace{-3mm}
\end{figure}

{\flushleft {\bf Segmentation}}
We use the pre-defined training set of the DAVIS benchmark \cite{Perazzi2016} to train the segmentation branch.
Since this training set is relatively small, we adopt affine transformations (i.e., shifting, rotation, flip) to generate one thousands samples for each frame.
Since the flow branch requires two adjacent frames as the input, each affine transformation is carried out through the entire sequence to maintain the inter-frame (temporal) consistency during training (see Figure \ref{fig:data_augmentation} for an example).

{\flushleft {\bf Optical Flow.}}
The flow data during offline training step is generated as the approach described for segmentation.
However, when training the online model using the first frame of a test set video, we have no access to its next frame.
To solve this problem, we present an optical flow data augmentation strategy.
First, we augment the first frame with the transformed method used in segmentation.
Then, based on each image and its object mask, we simulate an object movement by slightly deforming the foreground object region to generate a synthesized "next frame".
Since we only focus on the specific object at this online stage, the missing area caused by the object movement can be treated as occlusions and is left as empty (black) area.
We find this synthesized strategy is effective for training without harming the network property (see Figure \ref{fig:data_augmentation} for an example).
%
\section{Experimental Results}

We present the main experimental results with comparisons to the state-of-the-art video object segmentation and optical flow methods.
More results and videos can be found in the supplementary material.
The code and model are available at \url{https://github.com/JingchunCheng/SegFlow}.
%
%
\subsection{Dataset and Evaluation Metrics}
\label{sec:dataset}

The DAVIS benchmark \cite{Perazzi2016} is a recently-released high-quality video object segmentation dataset that consists of 50 sequences and 3455 annotated frames of real-world moving objects.
The videos in DAVIS are also categorized according to various attributes, such as background clutter (BC), deformation (DEF), motion blur (MB), fast motion (FM), low resolution (LR), occlusion (OCC), out-of-view (OV), scale-variation (SV), appearance change (AC), edge ambiguity (EA), camera shake (CS), heterogeneous objects (HO), interesting objects (IO), dynamic background (DB), shape complexity (SC), as shown in Figure \ref{fig:DAVIS_attibutes}.
We use the pre-defined training set to optimize our framework and its validation set to evaluate the segmentation quality.

For optical flow, we first use the MPI Sintel dataset \cite{Butler:ECCV:2012} that contains 1041 pairs of images in synthesized scenes, with a $Clean$ version containing images without motion blur and atmospheric effects,
and a $Final$ version of images with complicated environment variables.
Second, we use the KITTI dataset \cite{Geiger2012CVPR}, which has 389 pairs of flow images for real-world driving scenes.
Finally, we use the Scene Flow dataset \cite{MIFDB16}, which is a large-scale synthesized dataset recently established for flow estimation.
Considering the realism, we use two subsets, Monkaa and Driving, where Monkaa has a collection of 24 video sequences with more than 34000 annotations for optical flow, and Driving has 8 videos with around 17000 annotations.
Similar to Sintel, Driving and Monkaa both provide two versions: $Clean$ with clear images and $Final$ with more realistic ones.

Since the exact training and test sets are not specified in the Scene Flow dataset, we split our own sets for comparisons (training and validation sets do not intersect).
For Monkaa, we use three videos ($eating\_\times2$, $flower\_storm\_\times2$, $lonetree\_\times2$) as the validation set, and use the rest of 21 sequences for training.
For Driving, 7 videos are selected for training, and use the one with the attribute of $15mm\_focallength$, $scene\_forwards$ and $fast$ for testing.
Note that, every video in both Monkaa and Driving has two views of $left$ and $right$, which results in 63400 training and 5720 validation pairs on Monkaa,
and 32744 training and 2392 validation pairs on Driving.

To evaluate the segmentation quality, we use three measures (evaluation code from DAVIS website \cite{Perazzi2016}): region similarity $J$, contour accuracy $F$ and temporal stability $T$.
For optical flow, we compute the average endpoint error from every pixel for evaluation.

\subsection{Ablation Study on Segmentation}
\label{sec:ablation}
To analyze the necessity and importance of each step in the proposed framework, we carry out extensive ablation studies on DAVIS, and summarize the results in Table \ref{tab:ablation}.
We validate our method by comparing the proposed \textit{SegFlow} to the ones without online training ({\bf -ol}), iterative training ({\bf -it}), offline training ({\bf -of}) and flow branch ({\bf -flo}).
The detailed settings are as follows:

\noindent
{\bf -ol}: only uses the offline training without the supervised information in the first frame, which is categorized as unsupervised video object segmentation.

\noindent
{\bf -it}: only trains the model once for each of the segmentation and optical flow branches.

\noindent
{\bf -of}: trains the model directly on the testing video with the object mask in the first frame and its augmentations.

\noindent
{\bf -flo}: only uses the segmentation branch without the feature propagation from the flow network.

Table \ref{tab:ablation} shows that the offline training plays an important role in generating better results, improving the $J mean$ by 21\%.
It demonstrates that the network needs a generic model to discover moving objects before online finetuning.
The combined online and iterative strategy also improve the overall $J mean$ by 7.9\%.
Compared to the model without using the flow branch, our joint model not only improves the $J mean$ but also produces smooth results temporally, resulting in a significant improvement in $T mean$ by 5.6\%.

We evaluate the effectiveness of our data augmentation steps in Table \ref{tab:ablation}.
Without the data augmentation for segmentation ({\bf -sda}) and augmented flow data ({\bf -fda}), the performance both degrades in terms of $J mean$.
In addition, the $T mean$ is worse without augmenting flow data ({\bf -fda}), which shows the importance of the synthesized data described in Section \ref{sec:DA}.
%
%
\begin{table}[t]\scriptsize
	\caption{Ablation study on the DAVIS validation set.
		We show comparisons of the proposed \textit{SegFlow} model with different components removed, i.e., online-training (ol), offline-training (of), iterative learning (it), flow data augmentation (fda), optical flow branch (flo) and segmentation data augmentation (sda).}
	\vspace{-3mm}
	\begin{center}
		\small
		\centering
		\renewcommand{\arraystretch}{1.5}
		\setlength{\tabcolsep}{2pt}
		\begin{tabular}{|l|c|c|c|c|c|c|c|c|}
			\hline
			\vspace{-2mm}
			Method                & Ours    & -ol        &      -of  &   -ol        & -fda     &   -flo &     -flo&   -flo \\
			\vspace{-2mm}
			&                                &            &           &     -it      &     &            &      -ol&   -ol  \\
			&&&&&&&                                                                                                &   -sda \\
			\hline\hline
			J Mean $\uparrow$         &\bf 0.748       & 0.674      &    0.538  &    0.669&0.739&      0.724 &       0.654 &    0.606 \\
			J Recall $\uparrow$               & \bf 0.900      & 0.814      &     0.575 &    0.803&0.891&      0.882 &        0.787&    0.677 \\
			J Decay $\downarrow$                & 0.137          &0.062       &     0.227 &    0.005&0.124&      0.119 &       0.021 &\bf 0.006 \\
			\hline
			F Mean $\uparrow$                 &\bf 0.745       & 0.667      &     0.515 &    0.658&0.741&      0.735 &        0.640&    0.604 \\
			F Recall $\uparrow$               &\bf 0.853       & 0.771      &      0.540&    0.765&0.839&      0.841 &        0.750&    0.717 \\
			F Decay $\downarrow$                &  0.136         &  0.051     &     0.251 &    0.043&0.122&      0.132 &        0.017&\bf 0.001\\
			\hline
			T Mean $\downarrow$                 &\bf 0.194       & 0.276      &      0.302&    0.279&0.225&      0.250 &       0.354 &    0.335 \\
			\hline
		\end{tabular}
	\end{center}
	
	\label{tab:ablation}
	\vspace{-5mm}
\end{table}

\begin{table*}[t]\footnotesize
	\caption{Overall results on the DAVIS validation set with the comparisons to unsupervised and semi-supervised methods.}
	\vspace{1mm}
	\begin{center}
		\small
		\centering
		\renewcommand{\arraystretch}{1.2}
		\setlength{\tabcolsep}{3.5pt}
		\begin{tabular}{|ll|ccccccc|ccccccc|}
			\hline
			&&\multicolumn{7}{|c|} {\textcolor{blue}{Semi-Supervised}}            &\multicolumn{7}{|c|} {\textcolor{blue}{Unsupervised}}\\
			\multicolumn{2}{|c|}{Measure}
			&Ours$^2$&Ours    &Ours-flo     & OSVOS     &MSK         &OFL   &BVS      &   Ours-ol  &  Ours-flo-ol  & OSVOS &       FST &       CVOS&    KEY   &  NLC        \\
			\hline\hline
			J&Mean $\uparrow$ &   0.761 &0.748   &0.724        &{\bf 0.798}& 0.797      &0.680 &0.600       & {\bf 0.674}&          0.654&      0.525 &      0.558&  0.482    &  0.498    &    0.551   \\
			& Recall $\uparrow$&     0.906  &0.900  &0.882        &{\bf 0.936}& 0.931      &0.756 & 0.669       & {\bf 0.814}&          0.787&      0.577 &      0.649&   0.540   &   0.591   &   0.558    \\
			& Decay $\downarrow$&0.121   &0.137    &0.119        &  0.149    & {\bf 0.089}&0.264 &0.289      &       0.062&          0.021&{\bf -0.019}&     0.000&    0.105   &    0.141  &    0.126   \\
			\hline
			F & Mean \,$\uparrow$&  0.760  &0.745   &0.735        &{\bf 0.806}& 0.754      &0.634 &0.588       & {\bf 0.667}&          0.640&      0.477 &      0.511&      0.447 &     0.427&    0.523   \\
			& Recall $\uparrow$&  0.855   &0.853    &0.842        &{\bf 0.926}& 0.871      &0.704 &0.679       & {\bf 0.771}&          0.750&      0.479 &      0.516&      0.526 &    0.375 &    0.519   \\
			& Decay $\downarrow$&  0.104   &0.136   &0.132        &     0.150 &{\bf 0.090}  &0.272 &0.213       &       0.051&          0.017& {\bf 0.006}&      0.029&    0.117   &    0.106 &    0.114   \\
			\hline
			T&Mean $\downarrow$
			&{\bf 0.182}  &0.194 &0.250        &0.376        & 0.211      &0.217 &0.345       &       0.276&          0.354&      0.538 &     0.343 & {\bf 0.244} &    0.252&   0.414  \\
			\hline
		\end{tabular}
	\end{center}
	\label{tab:method_comparision}
	\vspace{-3mm}
\end{table*}
\subsection{Segmentation Results}
\label{sec:segmentatoin_results}
Table \ref{tab:method_comparision} shows segmentation results on the DAVIS validation set.
We improve the performance by considering the prediction of the image and its flipping one, and averaging both outputs to obtain the final result, where we refer to as Ours$^2$. Without adding much computational cost, we further boost the $Jmean$ with 1.3\%.
We compare the proposed \textit{SegFlow} model with state-of-the-art approaches, including
unsupervised algorithms (FST \cite{Papazoglou_ICCV_2013}, CVOS \cite{taylor2015causal}, KEY \cite{Lee_ICCV_2011}, NLC \cite{faktor2014video}), and semi-supervised methods (OVOS \cite{caelles2016one}, MSK \cite{khoreva2016learning}, OFL \cite{tsai2016video}, BVS \cite{marki2016bilateral}).

Among unsupervised algorithms, our \textit{SegFlow} model with or without the flow branch both performs favorably against other methods with a significant improvement (more than 10\% in $J mean$).
For semi-supervised methods, our model performs competitively against OSVOS \cite{caelles2016one} and MSK \cite{khoreva2016learning}, where their methods require additional inputs (i.e., superpixels in OSVOS and optical flow in MSK\footnote{With image only as the input, the $J mean$ of MSK \cite{khoreva2016learning} on the DAVIS validation set is $69.8$, which is much lower than ours as $74.8$.} with CRF refinement) to achieve higher performance,
while our method only needs images as inputs.
Furthermore, we show consistent improvements over the model without the flow branch, especially in the temporal accuracy ($T mean$), which demonstrates that feature representations learned from the flow network help the segmentation.

\begin{figure}
	\begin{center}
		\includegraphics[width=1.0\linewidth]{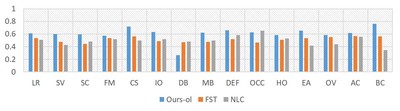}
	\end{center}
	\vspace{-3mm}
	\caption{Attribute based evaluation on the DAVIS validation set using $Jmean$ compared with unsupervised methods.}
	\label{fig:DAVIS_attibutes}
\end{figure}

\begin{figure*}[t]
	\begin{center}
		\includegraphics[width=1.0\linewidth]{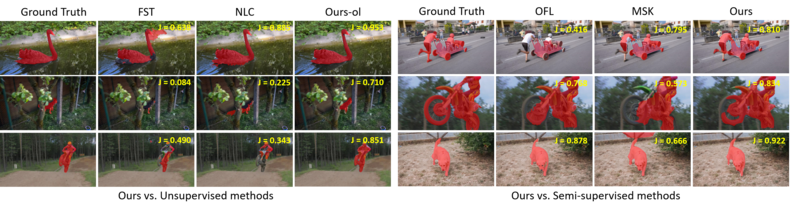}
	\end{center}
	\vspace{-5mm}
	\caption{Qualitative results on the DAVIS validation set with comparisons to unsupervised and semi-supervised algorithms.		
	}
	\label{fig:segmentation_comparison}
	\vspace{-1mm}
\end{figure*}

Figure \ref{fig:DAVIS_attibutes} shows the attributes-based performance ($J mean$) for different methods.
Our unsupervised method (offline training) performs well on all the attributes except for Dynamic Background (DB).
One possible reason is that motion representations generated from the flow branch may not be accurate due to the complexity in the background.
Figure \ref{fig:segmentation_comparison} presents some example results for segmentation.
With the flow branch jointly trained with segmentation, the model is able to recover the missing area of the object that is clearly a complete region from the flow estimation.
A full comparison per sequence and more results are provided in the supplementary material.
\begin{table}[!t]\footnotesize
	\caption{Average endpoint errors for optical flow. FlowNetS+ft denotes the results presented in \cite{fischer2015flownet}. FlowNetS+ft$^\ast$ denotes FlowNetS trained with the same data as SegFlow+ft.
		}
	\vspace{-3mm}
	\begin{center}
		\small
		\centering
		\renewcommand{\arraystretch}{1.2}
		\setlength{\tabcolsep}{1.5pt}
		\begin{tabular}{|l|cc|cc|c|cc|}
			\hline
			Method                                   & \multicolumn{2}{c|}{Sintel $Clean$}  & \multicolumn{2}{c|}{Sintel $Final$} & Chairs&\multicolumn{2}{c|}{KITTI}\\
			
			&train     & test                      & train    & test                     & test &train & test\\
			\hline\hline
			EpicFlow \cite{revaud2015epicflow}       &2.40      &4.12      &3.70      &6.29                      & 2.94  &3.47 &3.8\\
			DeepFlow \cite{weinzaepfel2013deepflow}  &3.31      &5.38      &4.56      &7.21                      & 3.53  &4.58 &5.8\\
			EPPM \cite{Bao_CVPR_2014}                &-         &6.49      &-         &8.38                      & -     & -   &9.2\\
			LDOF \cite{Brox_PAMI_2011}               &4.29      &7.56      &6.42      &9.12                      &3.47  &13.73 &12.4\\
			\hline
			FlowNetS \cite{fischer2015flownet}       &4.50      &7.42      &5.45      &8.43                      & 2.71 &8.26 & -\\
			FlowNetS+ft                            &2.97      &6.16      &4.07      &7.22                      &3.03 	&6.07 &7.6\\
		    FlowNetS+ft$^\ast$                           &3.31      &7.89      &4.26      &8.50                      & -    &7.37 &8.7\\            
			SegFlow+ft                               &2.50      &7.45      &2.61      &7.87                      &2.83  &4.40 &7.1\\
			\hline
		\end{tabular}
	\end{center}
	
	\label{tab:flow_methods_comparision}
	\vspace{-5mm}
\end{table}
\begin{figure*}[t]
	\begin{center}
		\includegraphics[width=1.0\linewidth]{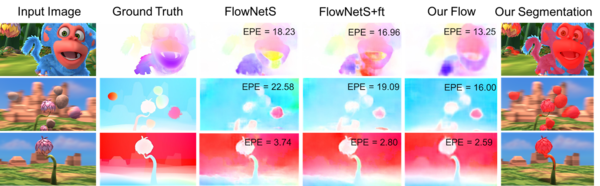}
	\end{center}
	\vspace{-3mm}
	\caption{
		For each input image, we show the optical flow results of the baseline FlowNetS \cite{fischer2015flownet}, fine-tuned FlowNetS and our \textit{SegFlow} model on the Scene Flow dataset.
		Our method produces outputs with lower endpoint error, especially with the visual improvement within the object, in which the flow is smoother than the other methods due to the guidance from the segmentation network.
	}
	\label{fig:flow_comparison}
	\vspace{-1mm}
\end{figure*}
\begin{table*}[!h]\footnotesize
	\caption{Average endpoint errors on the Scene Flow dataset. The evaluations for train and val on the Monkaa and Driving datasets use both $forward$ and $backward$ samples, while evaluations on train+val use $forward$ ones with the comparison as reported in \cite{MIFDB16}.}
	\vspace{-2mm}
	\begin{center}
		\small
		\centering
		\renewcommand{\arraystretch}{1.2}
		\setlength{\tabcolsep}{3pt}
		\begin{tabular}{|l|ccc|cc|ccc|cc|}
			\hline
			Method                                   & \multicolumn{3}{c|}{Monkaa $Clean$}  & \multicolumn{2}{c|}{Monkaa $Final$} &\multicolumn{3}{c|}{Driving $Clean$} & \multicolumn{2}{c|}{Driving $Final$}\\
			
			& train   & val    &train+val    & train   & val          & train   & val     &train+val     & train   & val \\
			\hline\hline
			SceneFlowNet \cite{MIFDB16}              &-        &-       &6.54         &-        &-             &-        &-        &23.53         &-        &-\\			
			FlowNetS \cite{fischer2015flownet}       &5.60     &10.51   &6.15         &5.48     &10.47         &13.29    &66.93    &23.90         &13.14    &67.15 \\
			FlowNetS+ft                              &4.93     &8.40    &5.01         &4.37     &8.44          &10.31    &52.67    &18.22         &10.38    &52.20\\
			SegFlow+ft                               &\bf 4.06 &\bf 7.94&\bf 4.49    &\bf 3.78 &\bf 7.90       &\bf 9.17 &\bf 37.91&\bf 14.35     &\bf 9.41 &\bf 37.93\\
			\hline
		\end{tabular}
	\end{center}
	\label{tab:flow_methods_comparision_scene_flow}
	\vspace{-5mm}
\end{table*}
\subsection{Optical Flow Results}
\label{sec:flow_result}
Table \ref{tab:flow_methods_comparision} and Table \ref{tab:flow_methods_comparision_scene_flow} show the average endpoint error of
the proposed \textit{SegFlow} model and the comparisons to other state-of-the-art methods,
including our baseline model (FlowNetS) used in the flow branch.
%
To validate the effectiveness of our joint training scheme, we use the pre-trained FlowNetS on the Flying Chair dataset \cite{fischer2015flownet} as the baseline, and finetune on the target dataset using the FlowNetS and our \textit{SegFlow} model for comparisons.

We note that the data layer used in \cite{fischer2015flownet} is specifically designed for FlowNetS, and thus we cannot directly apply it to our model.
Hence we report performance using various training data, where FlowNetS+ft denotes the results reported in \cite{fischer2015flownet} and FlowNetS+ft$^\ast$ denotes the model finetuned with the same training data as used in \textit{SegFlow}. 
As a result, we show that our \textit{SegFlow} model consistently improves endpoint errors against the results of FlowNetS+ft$^\ast$, which validates the benefit of incorporating the information from the segmentation branch. 
On KITTI, SegFlow without any data augmentation even outperforms FlowNetS+ft that uses extensive data augmentation. 
However, we observe that our model slightly overfits to the data on Sintel, due to the need of data augmentation on a much smaller dataset than the others.

In Table \ref{tab:flow_methods_comparision_scene_flow}, we also compare the results with SceneFlowNet \cite{MIFDB16} on the training and validation sets of Monkaa and Driving, and show that our method performs favorably against it.
Figure \ref{fig:flow_comparison} shows some visual comparisons of optical flow.
Intuitively, the segmentation provides the information to guide the flow network to estimate the output that aligns with the segmentation output (e.g., the flow within the segmentation is smooth and complete).
More results and analysis are provided in the supplementary material.

\subsection{Runtime Analysis}
\label{sec:runtime}
For the model trained offline, the proposed \textit{SegFlow} predicts two outputs (segmentation and optical flow) simultaneously in 0.3 seconds per frame on a Titan X GPU with 12 GB memory.
When taking the online training step into account, our system runs at 7.9 seconds per frame averaged over the DAVIS validation set.
Compared to other methods such as OFL (30 seconds per frame for optical flow generation and 90 seconds per frame for optimization), MSK (12 seconds per frame) and OSVOS (more than 10 seconds per frame at its best performance), our method is faster and can output an additional result of optical flow.
%
\vspace{-3mm}
\section{Concluding Remarks}
This paper proposes an end-to-end trainable network \textit{SegFlow} for joint optimization of video object segmentation and optical flow estimation.
We demonstrate that with this joint structure, both segmentation and optical flow can be improved via bi-directional feature propagations.
To train the joint model, we relax the constraint of a large dataset that requires both foreground segmentation and optical flow ground truths
by developing an iterative training strategy.
We validate the effectiveness of our joint training scheme through extensive ablation studies and show that our method performs favorably on both the video object segmentation and optical flow tasks.
The proposed model can be easily adapted to other architectures and can be used for joint training other co-related tasks.
%

\vspace{1mm}
\noindent
{\bf Acknowledgments.}
This work is supported in part by the NSF CAREER Grant \#1149783 and NSF IIS Grant \#1152576.
%

\clearpage
{\small
	\bibliographystyle{ieee}
	\bibliography{mybib}
}

\end{document}